# An Ensemble of Adaptive Neuro-Fuzzy Kohonen Networks for Online Data Stream Fuzzy Clustering

**Zhengbing Hu**
School of Educational Information Technology, Central China Normal University, Wuhan, China
Email: hzb@mail.ccnu.edu.cn

**Yevgeniy V. Bodyanskiy**
Kharkiv National University of Radio Electronics, Kharkiv, Ukraine,
Email: yevgeniy.bodyanskiy@nure.ua

**Oleksii K. Tyshchenko and Olena O. Boiko**
Kharkiv National University of Radio Electronics, Kharkiv, Ukraine,
Email: lehatish@gmail.com, olena.boiko@ukr.net

*Abstract*— A new approach to data stream clustering with the help of an ensemble of adaptive neuro-fuzzy systems is proposed. The proposed ensemble is formed with adaptive neuro-fuzzy self-organizing Kohonen maps in a parallel processing mode. Their learning procedure is carried out with different parameters that define a nature of cluster borders' blurriness. Clusters' quality is estimated in an online mode with the help of a modified partition coefficient which is calculated in a recurrent form. A final result is chosen by the best neuro-fuzzy self-organizing Kohonen map.

## I. Introduction

Multidimensional data clustering is common in Data Mining tasks. Such application areas as Text Mining and Web Mining have become really widespread lately. A traditional approach to solving this sort of tasks assumes that each vector of a processed sequence may only belong to a single class. Although it's a more natural case when each specific observation may be attributed to several classes at the same time with different membership levels.

This situation is a subject under study for fuzzy cluster analysis [1, 2]. In this approach, the most effective and simplest methods are probabilistic fuzzy clustering procedures based on optimization of some objective functions. Initial data for a fuzzy clustering problem is a sample of observations which consists of $N$ $(m \times 1)$ – dimensional feature vectors $X = \{x(1), x(2), \ldots, x(k), \ldots x(N)\} \subset R^n$, and a result of this clustering procedure is a partition of the initial data set into $m$ overlapping classes with some membership levels $0 < u_j(k) < 1$ of the $k$ – th feature vector to the $j$ – th cluster, $j = 1, 2, \ldots, m$. Thus, the overwhelming majority of the well-known fuzzy clustering algorithms is designated for a batch mode processing which means that a sample volume $N$ can't be changed while the data are processed.

There's a wide class of tasks to be solved only with the help of the Data Stream Mining [3-16] approach when data are fed and processed in an online mode. This task is rather typical for Web Mining when information is fed in a real time mode directly from the Internet.

Self-organizing maps (SOMs) by Kohonen proved its efficiency in clustering tasks. Their efficiency is defined by their computational simplicity and their ability to work in a real time mode for sequential data processing. These neural networks are learnt with the help of self-learning procedures based on the principles "Winner takes all" (WTA) and "Winner takes more" (WTM). It's previously assumed that a structure of processed data implies that formed clusters don't mutually intersect which means that it's possible to build a separating hyper-surface which clearly distinguish different classes during a learning procedure of a neural network.

Recurrent modifications of the fuzzy clustering algorithms (which make it possible to solve a task in an online mode) were introduced for sequential data processing in [17, 18]. It should be noted that the introduced procedures are structurally close to the Kohonen self-learning rule according to the principle «Winner Takes More». It allows introducing a so-called «fuzzy clustering Kohonen network» [19] which possesses a number of advantages comparing to a conventional self-organizing map.

The well-known and most commonly used fuzzy clustering algorithms can't be called fuzzy in the full sense, because their results are significantly defined by a value of a special parameter (also known as a fuzzifier $\beta$ which is chosen empirically). A case when $\beta$ belongs to an interval from 1 to $\infty$ corresponds to a transition from crisp borders $(\beta \to 1)$, which are obtained with the help of the K-means procedure, to their complete blurriness $(\beta \to \infty)$, when all observations belong to all clusters with the same membership level. We should note that $\beta = 2$ in most cases that corresponds to the fuzzy C-means procedure (FCM) by Bezdek [20].

There may be a situation while processing real-world data when one object belongs to different classes at the same time and these classes mutually intersect (overlap). Conventional SOMs don't take into consideration this occasion, but this problem can be considered with the help of fuzzy clustering techniques.

The remainder of this paper is organized as follows: Section 2 describes fuzzy clustering techniques with a variable fuzzifier. Section 3 describes an ensemble's architecture of adaptive neuro-fuzzy Kohonen networks. Section 4 gives some details on possibilistic fuzzy clustering with a variable fuzzifier. Section 5 presents a real-world application to be solved with the help of the proposed fuzzy clustering approach. Conclusions and future work are given in the final section.

## II. Fuzzy Clustering with a variable fuzzifier

Algorithms based on goal functions are considered to be strict from a mathematical point of view among all clustering procedures. They solve a task of their optimization under different a priori assumptions. The most commonly used procedure in this situation is the probabilistic approach which is based on a goal function's minimization

$$E(u_j, c_j) = \sum_{k=1}^{N} \sum_{j=1}^{m} u_j^\beta(k) \|x(k) - c_j\|^2 \qquad (1)$$

under constraints

$$\sum_{j=1}^{m} u_j(k) = 1, \qquad (2)$$

$$0 \leq \sum_{k=1}^{N} u_j(k) \leq N \qquad (3)$$

where $u_j(k) \in [0,1]$ is a membership level of a vector $x(k)$ to the $j$-th class, $c_j$ is a prototype of the $j$-th cluster, $\beta$ is a non-negative fuzzification parameter (a fuzzifier) which actually determines a level of borders' blurriness between clusters, $k = 1, 2, \ldots, N$. A result of this clustering procedure is a $(N \times m)$-matrix $U = \{u_j(k)\}$ which is also called a fuzzy partition matrix.

We should notice that elements of the matrix $U$ due to the constraint (2) may be considered as probabilities that data vectors belong to some definite clusters. Because of this fact, procedures based on the minimization (1) are called probabilistic fuzzy clustering algorithms. A number of clusters $m$ is set beforehand and can't be changed during computation procedures.

Introducing the Lagrange function

$$L(u_j(k), c_j, \lambda(k)) = \sum_{k=1}^{N} \sum_{j=1}^{m} u_j^\beta(k) \|x(k) - c_j\|^2 + \\ + \sum_{k=1}^{N} \lambda(k) \left( \sum_{j=1}^{m} u_j(k) - 1 \right) \qquad (4)$$

(here $\lambda(k)$ is an undetermined Lagrange multiplier) and solving the Karush-Kuhn-Tucker system of equations, we can get a solution in the form

$$\begin{cases} u_j(k) = \dfrac{\left(\|x(k)-c_j\|^2\right)^{\frac{1}{1-\beta}}}{\sum_{l=1}^{m}\left(\|x(k)-c_l\|^2\right)^{\frac{1}{1-\beta}}}, \\ c_j = \dfrac{\sum_{k=1}^{N} u_j^{\beta}(k) x(k)}{\sum_{k=1}^{N} u_j^{\beta}(k)}, \\ \lambda(k) = -\left(\left(\sum_{l=1}^{m}\beta \|x(k)-c_l\|^2\right)^{\frac{1}{1-\beta}}\right)^{1-\beta}, \end{cases} \qquad (5)$$

which coincides with the Fuzzy C-Means algorithm (FCM) by J.Bezdek (when $\beta = 2$). And when $\beta \to 1$ its results are close to results of the well-known conventional crisp clustering algorithm (Hard K-Means, HKM).

As an alternative to procedures that use a fuzzifier $1 < \beta < \infty$, Klawonn and Hoeppner [21] offered an objective function for fuzzy probabilistic clustering

$$E(u_j, c_j) = \sum_{k=1}^{N}\sum_{j=1}^{m}\left(\alpha u_j^2(k) + (1-\alpha) u_j(k)\right)\|x(k)-c_j\|^2 \qquad (6)$$

with the constraints (2) and (3), where $0 < \alpha \leq 1$ is an adjustable parameter which defines a nature of the obtained solution.

Introducing the Lagrange function

$$L(u_j(k), c_j, \lambda(k)) = \sum_{k=1}^{N}\sum_{l=1}^{m}\left(\alpha u_j^2(k) + (1-\alpha) u_j(k)\right)\|x(k)-c_j\|^2 + \sum_{k=1}^{N}\lambda(k)\left(\sum_{j=1}^{m} u_j(k) - 1\right)$$

and solving the Karush-Kuhn-Tucker system of equations

$$\begin{cases} \dfrac{\partial L(u_j(k), c_j, \lambda(k))}{\partial u_j(k)} = (2\alpha u_j(k) + 1 - \alpha)\|x(k)-c_j\|^2 + \lambda(k) = 0, \\ \nabla_{c_j} L(u_j(k), c_j, \lambda(k)) = -\sum_{k=1}^{N} 2\left(\alpha u_j^2(k) + (1-\alpha) u_j(k)\right)(x(k)-c_j) = \vec{0}, \\ \dfrac{\partial L(u_j(k), c_j, \lambda(k))}{\partial \lambda(k)} = \sum_{j=1}^{m} u_j(k) - 1 = 0, \end{cases}$$

we come to a solution

$$\begin{cases} u_j(k) = -\dfrac{1-\alpha}{2\alpha} + \dfrac{1 + m\dfrac{1-\alpha}{2\alpha}}{\sum_{l=1}^{m}\dfrac{\|x(k)-c_j\|^2}{\|x(k)-c_l\|^2}}, \\ c_j = \dfrac{\sum_{k=1}^{N}\left(\alpha u_j^2(k) + (1-\alpha) u_j(k)\right) x(k)}{\sum_{k=1}^{N}\left(\alpha u_j^2(k) + (1-\alpha) u_j(k)\right)}. \end{cases} \qquad (7)$$

It's easy to notice that when $\alpha = 1$ this procedure coincides with FCM. Thus, the procedure (7) can't be used for solving Data Stream Mining tasks, because it can't process information in an online mode. Therefore, an adaptive modification of the expression (7) was introduced in [22]

$$\begin{cases} u_j(k+1) = -\frac{1-\alpha}{2\alpha} + \frac{1+m\frac{1-\alpha}{2\alpha}}{\sum_{l=1}^{m} \frac{\|x(k+1)-c_j(k)\|^2}{\|x(k+1)-c_l(k)\|^2}}, \\ c_j(k+1) = c_j(k) + \eta(k)\left(\alpha u_j^2(k+1) + (1-\alpha)u_j(k+1)\right)\left(x(k+1)-c_j(k)\right) \end{cases} \quad (8)$$

where $\eta(k)$ is a learning rate parameter. It's easy to notice that the second recurrent expression (8) is the Kohonen self-learning rule according to the principle «Winner Takes More» with a neighborhood function $\alpha u_j^2(k+1) + (1-\alpha)u_j(k+1)$.

### III. AN ENSEMBLE OF ADAPTIVE NEURO-FUZZY KOHONEN NETWORKS

Although a value of the parameter $\alpha$ in the formulas (7) and (8) lies in a much narrower range than a fuzzifier $\beta$, but there are currently no formal rules how to choose it and to tune it. Therefore, while solving a concrete task in a batch mode, this task is usually repeatedly solved with the help of the expression (7) with different $\alpha$ values (from a very small quantity to 1). It's clear that such an approach can't solve tasks effectively in an online mode. It might be expedient to use an idea of an ensemble of parallel working clustering procedures [23, 24] in this situation, where each clustering procedure works with a different from others $\alpha$ value. This ensemble can be easily implemented with the help of adaptive neuro-fuzzy Kohonen networks [25]. They are two-layer architectures where prototypes are clarified in the Kohonen competitive layer (which contains $m$ neurons $N_j^K$) and membership levels are calculated in the output layer (which contains $m$ neurons $N_j^M$).

There is an architecture of the two-layer adaptive neuro-fuzzy Kohonen network in Fig.1. There is also an ensemble formed by such networks in Fig.2.

A self-learning algorithm for the $p$-th ensemble's member $(p=1,2,\ldots,q)$ in an ensemble that contains $q$ neuro-fuzzy networks can be written down in the form

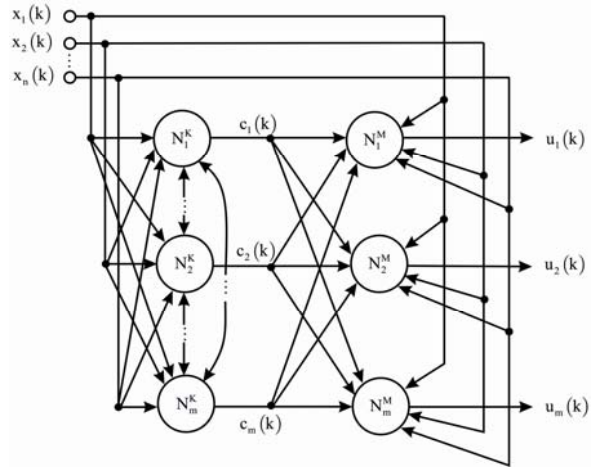

Fig.1. An adaptive neuro-fuzzy Kohonen network (FSOM)

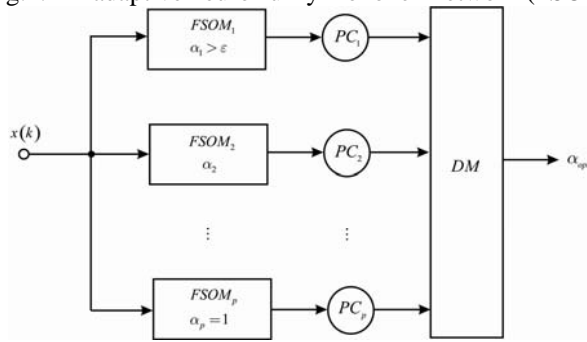

Fig.2. An architecture of an ensemble of adaptive neuro-fuzzy Kohonen networks

$$\begin{cases} c_{jp}(k+1) = c_{jp}(k) + \eta(k)\left(\alpha_p u_{jp}^2(k) + (1-\alpha_p)u_{jp}(k)\right)\left(x(k+1) - c_{jp}(k)\right), \\ u_{jp}(k+1) = -\frac{1-\alpha_p}{2\alpha_p} + \frac{1 + m\frac{1-\alpha_p}{2\alpha_p}}{\sum_{l=1}^{m} \frac{\|x(k+1) - c_{jp}(k)\|^2}{\|x(k+1) - c_{lp}(k)\|^2}} \end{cases} \quad (9)$$

wherein the Kohonen layer is tuned with the help of the first ratio (9), and the output layer calculates membership levels $u_{jp}(k+1)$ for each incoming observation $x(k+1)$.

Classification quality provided by each ensemble member may be estimated with the help of any fuzzy clustering index [2]. Wherein one of the simplest and most effective indexes is the so-called "partition coefficient" (PC) which is a mean value of squared membership levels of all observations to each cluster:

$$PC_p(k+1) = \frac{1}{k+1}\sum_{\tau=1}^{k+1}\sum_{j=1}^{m} u_{jp}^2(\tau). \quad (10)$$

This coefficient has a clear physical sense: the better clusters are expressed, the higher the value $PC_p$ is (a limit is $PC_p = 1$), its minimum ($PC_p = m^{-1}$) is reached if data belong to all clusters evenly. But this phenomenon is obviously a worthless solution. This coefficient is convenient in the framework of the proposed system, because it allows online calculation.

It should be noticed that the expression (10) is closely related to the traditional FCM. This coefficient must be modified for a considered case in the form

$$PC_p(k+1) = \frac{1}{k+1}\sum_{\tau=1}^{k+1}\sum_{j=1}^{m}\left(\alpha_p u_{jp}^2(\tau) + (1-\alpha_p)u_{jp}(\tau)\right)$$

which coincides with the expression (10) when $\alpha_p = 1$.

So, clustering a data stream that is fed in an online mode is solved with the help of parallel working adaptive neuro-fuzzy Kohonen networks which differ from each other only by a value of the parameter $\alpha_p$. Thus, the network's results are applied to a maximum value $PC_p(k+1)$ as a final result at any particular time point.

IV. POSSIBILISTIC FUZZY CLUSTERING WITH A VARIABLE FUZZIFIER

Basic problems that arise in fuzzy clustering have to do with constraints on a sum of membership values (they must be equal to 1). That's why algorithms that use the constraint (2) are called probabilistic fuzzy clustering algorithms. An existence of this constraint leads to the fact that an observation which doesn't belong to any class gets the same membership levels for all classes. One can avoid this drawback if the ideas of possibilistic fuzzy clustering are used [26]. These ideas are based on minimization of an objective function

$$E(u_j, c_j) = \sum_{k=1}^{N}\sum_{l=1}^{m} u_j^\beta(k)\|x(k) - c_j\|^2 + \sum_{j=1}^{m}\mu_j \sum_{k=1}^{N}\left(1 - u_j(k)\right)^\beta$$

where a scalar parameter $\mu_j > 0$ defines a distance where a membership level takes on a value of 0.5 which means that if

$$\|x(k) - c_j\|^2 = \mu_j$$

then

$$u_j(k) = 0.5.$$

Introducing an objective function [22, 25] similarly to (6)

$$E(u_j, c_j) = \sum_{k=1}^{N}\sum_{j=1}^{m}\left(\alpha u_j^2(k) + (1-\alpha)u_j(k)\right)\|x(k) - c_j\|^2 + \sum_{j=1}^{m}\mu_j \sum_{k=1}^{N}\left(\alpha(1 - u_j(k))\right)^2$$

and minimizing it by $u_j, c_j, \mu_j$, we come to a modified possibilistic procedure in the form

$$\begin{cases} u_j(k) = \dfrac{\left(\alpha\mu_j + (1-\alpha)\|x(k)-c_j\|^2\right) + \mu_j}{2\alpha\left(\|x(k)-c_j\|^2 + \mu_j\right)}, \\ \\ c_j = \dfrac{\sum_{k=1}^{N}\left(\alpha u_j^2(k) + (1-\alpha)u_j(k)\right)x(k)}{\sum_{k=1}^{N}\left(\alpha u_j^2(k) + (1-\alpha)u_j(k)\right)}, \\ \\ \mu_j = \dfrac{\sum_{k=1}^{N}\left(\alpha u_j^2(k) + (1-\alpha)u_j(k)\right)\|x(k)-c_j\|^2}{\sum_{k=1}^{N}\left(\alpha u_j^2(k) + (1-\alpha)u_j(k)\right)}. \end{cases} \quad (11)$$

It's interesting to note that this ratio for a prototype calculation in the formulas (7) and (11) completely coincides.

The expressions (11) can be written down for an adaptive case in the form

$$\begin{cases} u_j(k+1) = \dfrac{\left(\alpha\mu_j(k) + (1-\alpha)\|x(k+1)-c_j(k)\|^2\right) + \mu_j(k)}{2\alpha\left(\|x(k+1)-c_j(k)\|^2 + \mu_j(k)\right)}, \\ \\ c_j(k+1) = c_j(k) + \eta(k)\left(\alpha u_j^2(k+1) + (1-\alpha)u_j(k+1)\right)\left(x(k+1)-c_j(k)\right), \\ \\ \mu_j(k+1) = \left(\sum_{p=1}^{k+1}\left(\alpha u_j^2(p)+(1-\alpha)u_j(p)\right)\|x(p)-c_j(k+1)\|^2\right)\left(\sum_{p=1}^{k+1}\left(\alpha u_j^2(p)+(1-\alpha)u_j(p)\right)\right)^{-1}. \end{cases} \quad (12)$$

As one can see, the second recurrent ratio (12) is the WTM self-learning rule by Kohonen with the neighborhood function $\alpha u_j^2(k+1) + (1-\alpha)u_j(k+1)$.

Although the possibilistic algorithm (12) is a little more complicated from a computational point of view than the probabilistic procedure (8), its advantage is the fact that new clusters may be detected with the help of the possibilistic approach during online data processing. If a membership level of a new incoming observation $x(k+1)$ to all classes turns out to be lower than some predefined threshold then we can assume that there's a new $(m+1)-$th cluster and its initial prototype coordinates are

$$c_{m+1}(0) = x(k+1).$$

The algorithm (12) can be used as a learning procedure for the two-layer FSOM (Fig.1). Let's notice that an ensemble of clustering neural networks with the help of the possibilistic approach was introduced in [27, 28], but its unwieldiness impedes its usage while processing data streams. Simplicity of the ensemble's (Fig.2) numerical implementation makes it possible to process data in a real time mode.

## V. EXPERIMENTS

We have taken real-world data for our experiment. A data set describes students' knowledge status about the subject of Computer Science. The data set contains multivariate characteristics. It contains 302 instances with 5 attributes for each observation. Speaking of attribute information, the data set contains these attributes:
- STG (a degree of study time for goal object materials);
- SCG (a degree of a user's repetition number for goal object materials);
- STR (a degree of user's study time for related objects with a goal object);
- LPR (a user's exam performance for related objects with a goal object);
- PEG (a user's exam performance for goal objects).

We chose 3 attributes (STR, LPR, PEG) for the sake of visualization.

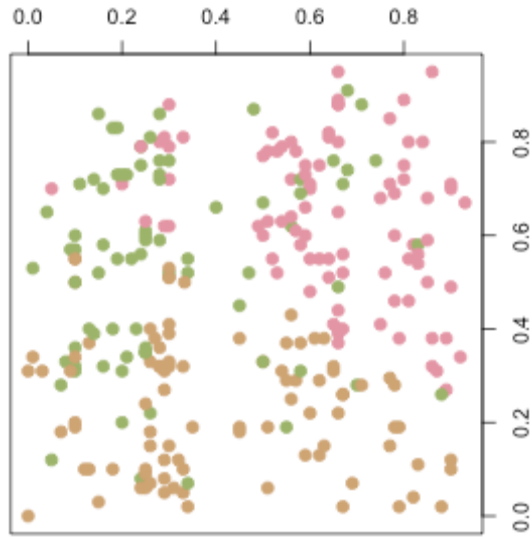

Fig.3. The Result of Fuzzy Clustering (STR/PEG)

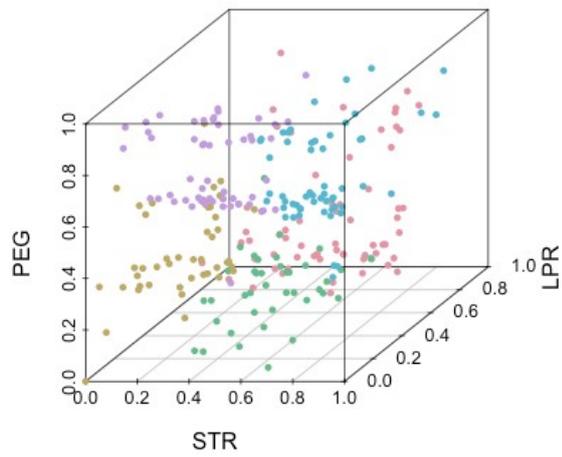

Fig.4. The Result of Fuzzy Clustering (PEG/ STR/LPR)

As it can be seen, the proposed method makes it possible to represent clusters in a rather compact form. A level of cluster overlapping is rather high (and it will keep on growing when a feature vector's dimensionality increases). An $\alpha$ parameter for the considered case belonged to an interval [0.3; 0.4].

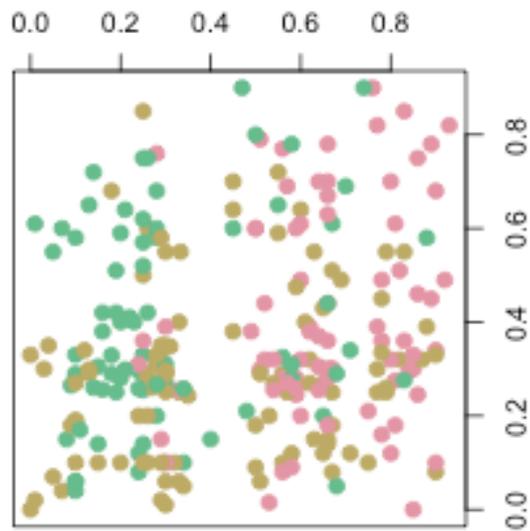

Fig.5. The Result of Fuzzy Clustering (SCG/PEG)

As it can be seen from Fig.3, there are 3 fuzzy clusters. Generally speaking, students' current knowledge level should be determined with the help of real values (STG, SCG, PEG, STR, LPR).

It should be noted that there are some points in Fig.3 which are not likely to be in those regions. For example, a point belongs to a "green" class although all neighbor points in that region belong to some other ("pink" or "yellow") class. Probably, this fact means that a student didn't actually spend much time on his exam preparation although he demonstrated a high level of performance (his PEG index is high). There may be another situation when a student spent much time on his preparation but his results aren't very good.

So, the clustering accuracy increases for fuzzy algorithms and worsens for crisp ones with the growth of a sample size; the clustering accuracy lessens with the growth of dimensionality of a feature space. Fuzzy procedures are preferable for those clustering tasks when every object belongs to several categories at the same time.

## VI. Conclusion

The method for online fuzzy clustering multidimensional data sequences to be processed in a real time mode is proposed. The task is solved with the help of an ensemble of adaptive neuro-fuzzy Kohonen networks which differ from each other by a parameter's value that accounts for fuzziness of the received results. The proposed procedure has rather simple computational implementation and makes it possible to organize parallel computing to accelerate the system's processing speed (because it can process data in an online mode and this fact is really important for Web Mining and Data Stream Mining). The results may be successfully used in a wide class of Data Stream Mining, Dynamic Data Mining, Temporal Data Mining tasks and especially in such applied areas as Web Mining, Text Mining, Medical Data Mining etc.

So, the proposed ensemble of adaptive neuro-fuzzy self-organizing Kohonen maps has proved its efficiency for online Data Stream fuzzy clustering. A number of experiments demonstrated a high effectiveness of the proposed neuro-fuzzy system especially under conditions of clusters' overlapping.

## Acknowledgment

The authors would like to thank anonymous reviewers for their careful reading of this paper and for their helpful comments.

This scientific work was supported by RAMECS and CCNU16A02015.